\documentclass[preprint, review,3p,times, 10pt]{elsarticle}

\usepackage{amssymb}

\usepackage{multirow}
 \usepackage[table,xcdraw]{xcolor}
\usepackage[utf8x]{inputenc}
\usepackage[T1]{fontenc}    % use 8-bit T1 fonts
\usepackage{booktabs}       % professional-quality tables
\usepackage{amsfonts}       % blackboard math symbols
\usepackage{nicefrac}       % compact symbols for 1/2, etc.
\usepackage{microtype}      % microtypography
\usepackage{lipsum}
\usepackage{graphicx}
\usepackage{amsmath}
\usepackage{verbatim}
\usepackage{float}
\usepackage[hyphens]{url}
\usepackage{hyperref}
\hypersetup{colorlinks=true,breaklinks=true}
\usepackage{caption}
\usepackage{subcaption}
\usepackage{makecell}
\usepackage[spanish,dutch,english]{babel}

\journal{arxiv}

\begin{document}

\begin{frontmatter}

%\title{COVIDHealth: A Benchmark Twitter Dataset for Classifying COVID-19 Related Discussions}
\title{COVIDHealth: A Benchmark Twitter Dataset and Machine Learning based Web Application for Classifying COVID-19 Discussions}

\author[inst1]{Mahathir Mohammad Bishal}
\ead{mahathirbishal@gmail.com}
\author[inst1]{Md. Rakibul Hassan Chowdory}
\ead{alvihasan361@gmail.com}
\author[inst2]{Anik Das}
\ead{x2021gmg@stfx.ca}
\author[inst3]{Muhammad Ashad Kabir\corref{corau}}
\ead{akabir@csu.edu.au}

\cortext[corau]{Corresponding author}
\affiliation[inst1]{organization={Department of Computer Science and Engineering, Chittagong University of Engineering and Technology},%Department and Organization
            %addressline={Address One}, 
            city={Chattogram},
            postcode={4349}, 
            %state={State One},
            country={Bangladesh}}

\affiliation[inst2]{organization={Department of Computer Science},
            addressline={St. Francis Xavier University}, 
            city={Antigonish},
            postcode={B2G 2W5}, 
            state={NS},
            country={Canada}}

\affiliation[inst3]{organization={Data Science Research Unit, School of Computing, Mathematics, and Engineering, Charles Sturt University},
        cite={Bathurst},
        postcode={2795},
        state={NSW},
        country={Australia}
        }

\begin{abstract}
%introduction, in this paper, title, contribution, method, result, discussion

%Twitter are nowadays one of the mainstream source for authentic and real-time information on the basis of different events.  However, extracting efficient information from twiter is a hard and challenging task because all the text data are unstructured. Researchers are trying to apply machine learning algorithm to the covid-19 dataset that is collected worldwide during this pandemic. 

The COVID-19 pandemic has had adverse effects on both physical and mental health. During this pandemic, numerous studies have focused on gaining insights into health-related perspectives from social media. In this study, our primary objective is to develop a machine learning-based web application for automatically classifying COVID-19-related discussions on social media. To achieve this, we label COVID-19-related Twitter data, provide benchmark classification results, and develop a web application. We collected data using the Twitter API and labeled a total of 6,667 tweets into five different classes: health risks, prevention, symptoms, transmission, and treatment. We extracted features using various feature extraction methods and applied them to seven different traditional machine learning algorithms, including Decision Tree, Random Forest, Stochastic Gradient Descent, Adaboost, K-Nearest Neighbour, Logistic Regression, and Linear SVC. Additionally, we used four deep learning algorithms: LSTM, CNN, RNN, and BERT, for classification. Overall, we achieved a maximum F1 score of 90.43\% with the CNN algorithm in deep learning. The Linear SVC algorithm exhibited the highest F1 score at 86.13\%, surpassing other traditional machine learning approaches. Our study not only contributes to the field of health-related data analysis but also provides a valuable resource in the form of a web-based tool for efficient data classification, which can aid in addressing public health challenges and increasing awareness during pandemics. We made the dataset and application publicly available, which can be downloaded from this link \url{https://github.com/Bishal16/COVID19-Health-Related-Data-Classification-Website}.
%In conclusion, deep learning algorithms perform better for our dataset than traditional machine learning algorithms. We are hopeful that our findings will be helpful to get more insight into COVID-19, as well as, assist medical and health organizations to fight against this pandemic. 

%We also compared the accuracy of  various machine learning and deep learning model. In our study we emphasized in the context of COVID-19 related data labelling according to the above mentioned classes.

%We have manually labelled the whole dataset and faced a huge challenge for doing this.
%As automated labelling approach is not applied here, we can only able to manually label  in our dataset. To overcome this problem we have collected tweet from twitter api and an open dataset. 

\end{abstract}


% \begin{highlights}
% \item A benchmark twitter dataset on COVIDHealth
% \item Crucial dataset sampling and feature extraction techniques
% \item Classified data using both traditional machine learning and deep learning
% \item Developed Web Application for Classifying COVID-19 Discussions
% \end{highlights}

% Highlights link (update here): https://docs.google.com/document/d/1mf12lCNWSIof2DXoyNP6X-t9D0HRvW7LSzj0qAFbC_8/edit


\end{frontmatter}



\section{Introduction}

Social media platforms, including Twitter, Facebook, Whatsapp, Weibo, and others, have evolved into powerful channels for real-time communication during natural disasters and disease outbreaks across the globe~\citep{reveilhac2022framing}.
These platforms have become primary mediums for individuals to communicate, share their experiences, and exchange thoughts~\citep{schillinger2020infodemics}. It holds the potential to serve as a valuable public health tool for scientists to promptly convey accurate information during pandemics, efficiently collecting reliable data~\citep{liu2020epic}. Today, researchers harness the wealth of unstructured data from social media to construct effective frameworks for healthcare applications~\citep{rufai2020world,lin2021social}.

Twitter, a microblogging and long-distance informal communication service, allows users to send ``tweets" limited to 280 characters. With over 368 million monthly active users worldwide~\citep{twitteruser}, it has become an essential platform for sharing ideas, data, and experimentation among medical experts for more than a decade~\citep{rosenberg2020twitter,arafat2022communication}.
It has emerged as a rapid and direct communication tool for disseminating COVID-19 information to the general public. People have turned to Twitter to share discussions related to the pandemic~\citep{chen2020COVID}.

User-generated content from social media platforms has gained substantial recognition for syndromic surveillance during global health emergencies, such as the 2009 H1N1 pandemic~\citep{achrekar2011predicting,chan2011using,chew2010pandemics,culotta2010towards,ginsberg2009detecting,lampos2010flu,lazer2014parable}, the 2014 Ebola outbreak~\citep{alicino2015assessing,jin2014misinformation,kalyanam2015facts,lu2015visualizing,odlum2015can,yom2015ebola}, the 2003 SARS epidemic~\citep{ehrenstein2006influenza}, and recently COVID-19 pandemic~\citep{shen2020using,mackey2020machine}.

COVID-19 presents a significant global health threat~\citep{chen2021high,das2021covid}, with particular severity observed in individuals with weakened immune systems, diabetes, or pre-existing conditions like lung or heart disease~\citep{WHOriskgroups}. This virus is highly contagious~\citep{shereen2020covid,guo2020origin}, and the analysis of its transmission identifies both direct modes, such as person-to-person contact~\citep{ghinai2020first}, and indirect pathways, including transmission via contaminated surfaces~\citep{world2020getting}. Notably, despite the pandemic's impact, there has been a notable absence of studies that focused on the analysis of health risks and transmission-related content of COVID-19 using social media data.

In this paper, we have meticulously classified health-related terms related to COVID-19 within Twitter data. We initiated by collecting tweet IDs from an open-source dataset, creating a tweet dataset from these tweet IDs, and categorizing the tweets into five distinct classes: health risks, prevention, symptoms, transmission, and treatment. Subsequently, we conducted preprocessing and applied data augmentation techniques to address dataset imbalances. We extracted features using three distinct methods, employing both traditional machine learning and deep learning approaches to classify the tweets. Our key contributions are as follows:

\begin{itemize}
    \item We have introduced a new COVID-19 Twitter dataset, facilitating the analysis and classification of COVID-19-related discussions based on five key categories: health risks, prevention, symptoms, transmission, and treatment.
   \item We have conducted a comprehensive empirical study using classical machine learning and deep learning approaches, offering a baseline classification performance for our dataset.
   \item To showcase the practical applicability, we have developed a web application prototype as a Chrome extension, utilizing the best model.
\end{itemize}

%This paper explores the methodology and findings related to the creation of this benchmark dataset, the classification techniques employed, and the development of a user-friendly web application for classifying COVID-19 discussions on Twitter.

%There are some COVID-19 related works which emphasize on labelling twitter data. But here we will classify our twitter data into 5 classes which is unique and haven't been done in other paper. We have emphasized in the context of COVID-19 related data labelling according to the above mentioned classes.
%How feature extraction, selection and various parameter affect the overall accuracy of algorithm are explained in this paper. We also present analysis between tradition machine learning and deep neural network algorithm. 
%According to the definition provided by CDC and WHO, our dataset is manually labelled by two annotators and crosschecked further.
%We added necessary graphs, table to compare various algorithm on the dataset.

The remaining sections of the paper are organized as follows: Section~\ref{sec:related work} discusses related works, and Section~\ref{sec:methodology} presents the workflow of our proposed methodology. Section~\ref{sec:Building the COVIDHealth Dataset} describes the details of collecting the dataset, labelling, preprocessing and the description of the dataset. Section~\ref{sec:Machine Learning Techniques for Text Classification} explains data sampling, feature extraction, and various classification methods. Section~\ref{sec:Experiments and Results} presents the outcomes of our experimental evaluation. After a brief discussion in Section~\ref{sec:Discussion}, Section~\ref{sec:Conclusion} concludes the paper by summarizing the key findings and future work in the field.

\section{Related Works}\label{sec:related work}

Online social media has played a pivotal role in infectious disease monitoring, prevention, and control for several years~\citep{hagg2018emerging}. Notably, one pertinent study focuses on the health domains of the 2014 Ebola and 2016 Zika outbreaks using Twitter data~\citep{khatua2019tale}. This research categorized tweets into five health perspectives: health risks, prevention, symptoms, transmission, and treatment for both Ebola and Zika outbreaks, followed by the application of pre-trained models, Word2Vec and GloVe, and an extra tree classifier for classification.

Preventive measures are crucial in curbing the spread of infectious diseases like COVID-19~\citep{omer2020preventive,ali2020covid}. However, only a limited number of studies have specifically addressed preventive measures, such as handwashing, social distancing, and face shields. Many of these studies have focused on opinion mining related to mask-wearing from tweets~\citep{cotfas2021unmasking,al2021public,he2021people}. An exception is the work by Doogan et al.~\citep{doogan2020public}, which concentrates on nonpharmaceutical interventions (NPIs) encompassing seven categories, including gathering restrictions, lockdowns, personal protection, social distancing, workplace closures, testing and tracing, and travel restrictions.

Identifying various combinations of symptoms is essential for characterizing infectious diseases~\citep{zhou2014human, emmett1998nonspecific}. Mackey et al.~\citep{mackey2020machine} have explored COVID-19 symptom self-reporting through bi-term topic modeling, an unsupervised machine learning approach applied to Twitter data. Shen et al.~\citep{shen2020using} have introduced a supervised machine-learning approach that leverages diagnosis reports and symptoms from Weibo posts to predict COVID-19 case counts. Other studies have focused on extracting prevalent symptoms from tweets~\citep{alanazi2020identifying} and utilizing symptom-related topics for sentiment analysis~\citep{xue2020public}.

Analysing treatments for COVID-19 is vital~\citep{khan2021covid}, and the long-term immunity and efficacy of approved COVID-19 vaccines are yet to be fully determined~\citep{tavilani2021covid}. Several studies have examined classes directly or indirectly related to treatment, encompassing public perception and opinion mining of COVID-19 vaccines~\citep{mir2021public,cotfas2021longest}, anti-vaccination sentiment identification~\citep{to2021applying}, detection of vaccine misinformation~\citep{weinzierl2021automatic}, and conspiracies~\citep{gerts2021thought} from social media. Nevertheless, these studies do not comprehensively cover all treatment-related topics of COVID-19, warranting more thorough investigation.

Multi-class classifications are inherently complex due to potential feature overlap between classes, in contrast to binary classification~\citep{koziarski2020combined}, and recognizing minority class features can be challenging~\citep{sahare2012review}. To date, no study has collectively focused on the mentioned classes in a single framework, with two of them lacking related works, and only a limited number of studies addressing the other three classes. The development of an automatic recognition system, such as a web application, for these significant topics holds potential to enhance the usability of classification outputs for both healthcare facilities and the general population.

 \section{Methodology}\label{sec:methodology}
% Figure ~\ref{workflow} depicts a high-level overview of our proposed methodology. To build the COVIDHealth dataset, in step 1, we collect tweets using the tweet IDs from COVID-19 Tweets dataset~\citep{lamsal2021design} and select relevant tweets using the keywords listed in Table~\ref{tab:Classification type with their definition}. Next, we label the selected tweets that we have collected. Two annotators labeled the tweets independently according to the definition provided in Table~\ref{tab:Classification type with their definition}. A third expert checked the labeling for any inconsistencies and resolved them through consensus. In the end, the final corpus is created. As text data is not directly usable for various machine learning and deep learning algorithms, we extracted the relevant features from the text after preprocessing in step 3. Here three different feature extraction techniques are used for this purpose. Following the feature extraction part, we feed those features into several machine learning and deep learning algorithms for the purpose of classification shown in steps 4.1 and step 4.2. Finally, in step 5, We investigated how well various machine learning and deep learning classifiers performed, and developed a web application prototype as a Chrome extension using the best model. The description of these steps is described in the following subsections.

Figure~\ref{workflow} provides a high-level overview of the proposed methodology employed in this study. The construction of the COVIDHealth dataset involved a series of key steps:
\begin{figure}[!htbp]
    \centering
    \includegraphics[width=16cm]{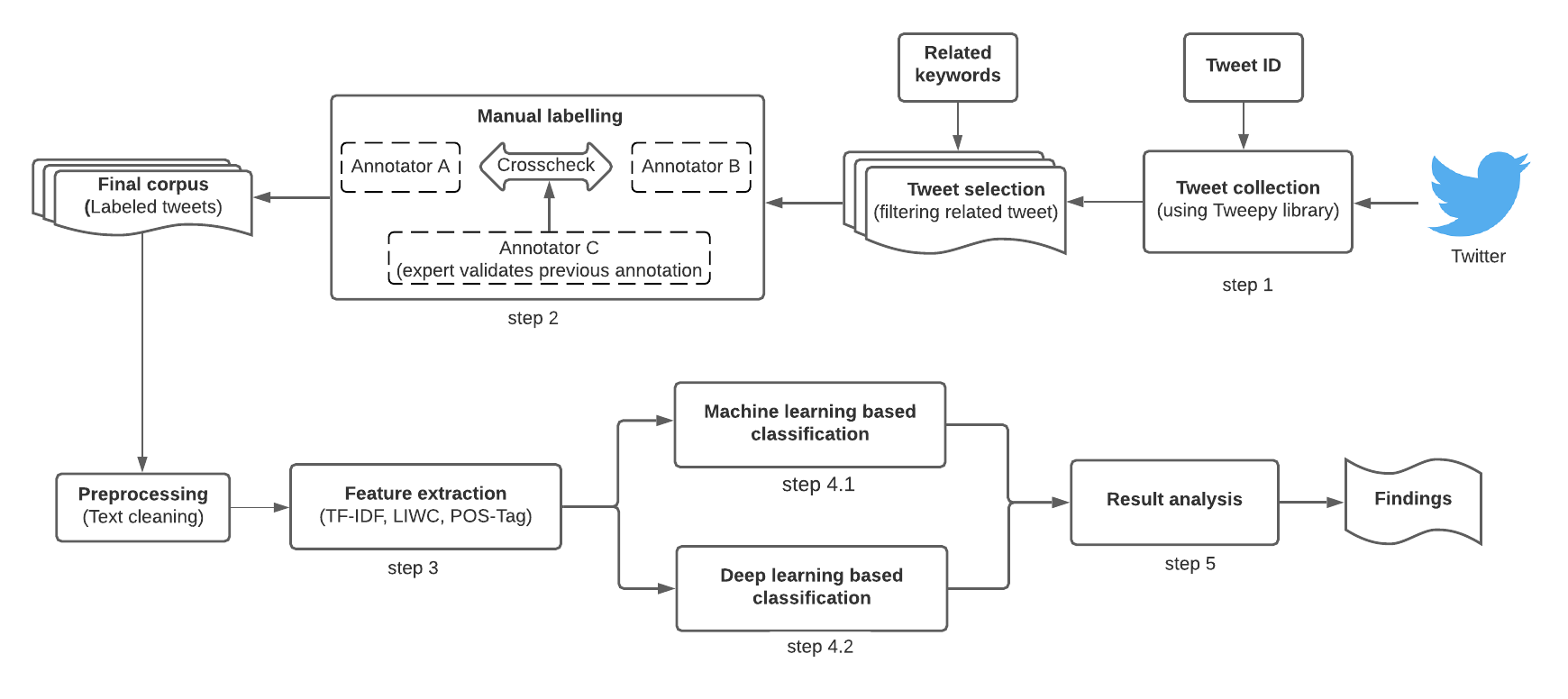}
    \caption{Workflow of our proposed methodology}
    \label{workflow}
\end{figure}

\textit{Data Collection}. In the first step, we gathered tweets by utilizing tweet IDs obtained from the COVID-19 Tweets dataset~\citep{lamsal2021design}. To ensure the relevance of the collected data, we employed predefined keywords as detailed in Section~\ref{sec:Building the COVIDHealth Dataset}.

\textit{Data Annotation}. Following data collection, we conducted a two-step annotation process. Initially, two independent annotators labeled the collected tweets based on the criteria outlined in Section~\ref{sec:Building the COVIDHealth Dataset}. Subsequently, a third expert meticulously reviewed the annotations, addressing any discrepancies through consensus to ensure the accuracy and consistency of the final dataset.

\textit{Preprocessing and Feature Extraction}. As raw text data is not directly compatible with various machine learning and deep learning algorithms, we performed text preprocessing in the third step. We implemented three distinct feature extraction techniques to derive meaningful features from the text data, outlined in Section~\ref{sec:featureextraction}.

\textit{Classification}. In steps 4.1 and 4.2, we fed the extracted features into a variety of machine learning and deep learning algorithms for classification purposes (described in Section~\ref{sec:classification}). This step encompassed the model training and evaluation phase to assess the performance of these algorithms.

\textit{Performance Evaluation and Application Development}. In the final step (step 5), we conducted a comprehensive analysis to evaluate the performance of the various machine learning and deep learning classifiers. Based on the results (reported in Section~\ref{sec:Experiments and Results}, we proceeded to develop a web application prototype as a Chrome extension, using the best-performing model (presented in Section~\ref{sec:application}).

The subsequent subsections of this article will provide a detailed description of each of these steps, offering insight into the methods and techniques employed in the creation of the COVIDHealth dataset and the subsequent analysis and application development.

%The methodology of this study includes dataset description, classification techniques and preparation of gold standards, data balancing and data analysis. In data analysis subsection traditional machine learning, feature extraction method and deep machine learning algorithm are discussed. Work flow of the whole process is given in Fig. \ref{workflow}. At first we have collected tweets using twitter API and tweepy library. Then the whole dataset is labelled by two annotator. One annotator crosschecked the dataset and corrected the mistake. Then the data dataset is preprocessed by removing mention, tagging, 'URL', irrelevant words, newline, repeated character, punctuation, stopwords etc. Then three feature extraction technique-LIWC, TF-IDF, POS tag are applied on the preprocessed dataset. After that the machine learning and deep learning models have been used with the extracted features. Seven machine learning and four deep learning algorithm have been used in this work. We have tested unlabeled dataset against those models to predict the correct class of the testing sample tweet. 

\section{Building the COVIDHealth Dataset}\label{sec:Building the COVIDHealth Dataset}
\subsection{Data collection and labelling}

We have used a publicly available dataset, CORONAVIRUS (COVID-19) TWEETS DATASET~\citep {lamsal2021design}, consisting of an extensive collection of 1,091,515,074 tweet IDs, and continuously expanding. The dataset was compiled by tracking over 90 distinct keywords and hashtags commonly associated with discussions about the COVID-19 pandemic. From this massive dataset, we focused on a specific time frame, encompassing data from August 05, 2020, to August 26, 2020, to meet our research objectives. As this dataset contains only tweet IDs, we have used the Twitter developer API to retrieve the corresponding tweets from Twitter. This retrieval process involved searching for tweet IDs and extracting the associated tweet texts, and it was implemented using the Twython library\footnote{\url{https://github.com/ryanmcgrath/twython}}. In total, we successfully collected 21,890 tweets during this data extraction phase.

Following guidelines set by the CDC and WHO, we categorized tweets into five distinct classes for classification: health risks, prevention, symptoms, transmission, and treatment, as detailed in Table~\ref{tab:Classification type with their definition}.
\begin{table}[!htb]
     \caption{Classification type with their definition}
      \centering
        \resizebox{1\textwidth}{!}{
        \begin{tabular}{ p{2.5cm}p{8cm}p{5.5cm}}
        \hline
        %\multicolumn{3}{|c|}{Country List} \\
        %\hline
        Class name & Situation in COVID-19 & Related keywords \\
        \hline\hline
        Health risks & People aged over their sixties or people with heart disease, lung problems, weak immune systems, or diabetes get more affected by COVID-19. & Lung disease, heart disease, diabetes, weak immunity, front line heroes\\
        
        %\hline
        Prevention & Avoiding close contact, covering sneezes and coughs, covering nose and mouth around others with face shields or covers, disinfecting and cleaning more often, washing hands frequently, and routine health monitoring. & Wash hands, homeschooling, close contact, cover mouth, cover nose, coughs, sneezes, clean and disinfect, face shields\\
        
        Symptoms & Common COVID-19 symptoms, e.g., cough or cold, congestion or runny nose, breathing issues, fever,  muscle or body aches, sore throat, diarrhea, nausea or vomiting, loss of taste or smell, headache, fatigue & Shortness of breath, cough, fever, chills, fatigue, vomiting, nausea, diarrhea, headache, sore throat\\
        %\hline
        Transmission & Person-to-person spread, the virus spreads easily between people, touching a surface or object that has the virus on it, spread between animals, people & Person-to-person spread, spreads easily between people, touching a surface, touching an object, spread between animals, spread between people  \\
        %\hline
        Treatment & Probable vaccine development and drugs used for COVID-19 treatment & Vaccine, drugs, paracetamol, herd immunity\\
        
        \hline
        
        \hline
         
        \end{tabular}
        }
      \label{tab:Classification type with their definition}
    \end{table}
Specifically, individuals aged over sixty, or those with pre-existing health conditions such as heart disease, lung problems, weakened immune systems, or diabetes, are at higher risk of severe COVID-19 complications. Therefore, tweets categorized as `health risks' pertain to the elevated risks associated with COVID-19 due to age or specific health conditions. `Prevention' related tweets encompass discussions on preventive and precautionary measures regarding the COVID-19 pandemic. Tweets discussing common COVID-19 symptoms, including cough, congestion, breathing issues, fever, body aches, and more, are classified as `symptoms' related tweets. Conversations pertaining to the spread of COVID-19 between individuals, between animals and humans, and contact with virus-contaminated objects or surfaces are categorized as `transmission' related tweets. Lastly, tweets indicating vaccine development and drugs used for COVID-19 treatment fall under the `treatment' related category.

We determined specific keywords for each of the five classes (health risks, prevention, symptoms, transmission, and treatment) based on the definitions provided by the CDC and WHO on their official websites. These definitions, along with their associated keywords, are detailed in Table~\ref{tab:Classification type with their definition}. For instance, the CDC and WHO indicate that individuals over the age of sixty with conditions like heart disease, lung problems, weak immune systems, or diabetes face a higher risk of severe COVID-19 complications. In accordance with this definition, we selected relevant keywords such as ``lung disease", ``heart disease", ``diabetes", ``weak immunity", and others to identify tweets related to health risks within the larger tweet dataset. This approach was consistently applied to define keywords for the remaining four classes. Subsequently, we filtered the initial dataset of 21,890 tweets to extract tweets relevant to our predefined classes, resulting in a total of 6,667 tweets based on the selected keywords.

To ensure the accuracy of our dataset, two separate annotators individually assigned the 6,667 tweets to the five classes. A third annotator, a natural language expert, meticulously cross-checked the dataset and provided necessary corrections. Subsequently, the two annotators resolved any discrepancies through mutual agreement, resulting in the final annotated dataset. 
Table~\ref{tab:No of tweet in each class} provides an overview of the distribution of tweets among the five classes, with 978, 2046, 1402, 802, and 1439 tweets annotated as `health risk', `prevention', `symptoms', `transmission', and `treatment', respectively (as presented in Table~\ref{tab:No of tweet in each class}). Additionally, Table~\ref{sample table} offers a selection of example tweets from each of the defined categories, providing insights into the nature of the annotated content.

%Then the final dataset is prepared. Table~\ref{tab:No of tweet in each class} represents the number of tweets in each class. Annotators have annotated 978, 2046, 1402, 802, and 1439 tweets as a health risk,  prevention, symptoms, transmission, and treatment respectively (presented in Table~\ref{tab:No of tweet in each class}). Table~\ref{sample table} shows a few examples of tweets from each category.
\begin{table}[!htb]
 \caption{Tweets distribution in dataset}
  \centering
    
    \begin{tabular}{lr}
    %\specialrule{.7pt}{0pt}{0pt}
    \hline
    Class     & Tweets count   \\
    \hline\hline
    %\specialrule{.9pt}{0pt}{0pt}
    
            Health risk & 978\\
            Prevention & 2,046\\
            Symptoms & 1,402\\
            Transmission & 802\\
            Treatment & 1,439\\
    \hline
    Total & 6,667\\
    \hline
    
    \hline
    %\specialrule{.7pt}{0pt}{0pt}
    
    \end{tabular}
  \label{tab:No of tweet in each class}
\end{table}

%but all these tweets are not relevant for this study. By using the related keywords mentioned in Table~\ref{tab:Classification type with their definition}, further .

\begin{table}[!htb]
     \caption{Example tweets from the dataset for each class}
      \centering
        \begin{tabular}{ p{2cm}p{14cm}}  \hline
        Class & Sample tweet \\ \hline\hline
        
        Health risk & COVID went after people with diabetes, obesity, high blood pressure. If you have  these, you're at higher risk of being infected. \\ \hline
        
        Prevention  & Use mask, avoid gathering, wash hand.\#stay\_home \#stay\_safe.\\
        \hline
        Symptoms   & Cough, sore throat, shortness breath, runny nose and loss of smell are primary symptoms of  covid19. \\ 
        \hline
        Transmission & Right now, about 100 students are in quarantine because of close contact with a positive \#Covid\_19 individual. \\
        \hline
        Treatment & Great reminder: until we have definitive pharmacological interventions for COVID, it's down to masks, ventilation, testing. \\ 
        \hline
        
        \hline
    \end{tabular}
\label{sample table}
\end{table}

% \begin{table}  [!htbp]
% \caption{Top-10 frequent words from tweets in the COVIDHEALTH dataset}
% \centering
% \begin{tabular}{l  r} 
% \hline
%         Keywords   & Count \\ \hline\hline
%         COVID      & 2,957 \\ 
%         vaccine    & 1,001 \\ 
%         people     & 684   \\ 
%         quarantine & 656   \\ 
%         lockdown   & 645   \\ 
%         blood      & 507   \\ 
%         high       & 444   \\ 
%         large      & 440   \\ 
%         pressure   & 410   \\ 
%         gathering  & 363   \\
% \hline

% \hline
% \end{tabular}
% \label{frequency table}
% \end{table}

\subsection{Dataset visualization}
\begin{figure}[!htb]
    \centering
    \includegraphics[width=9cm]{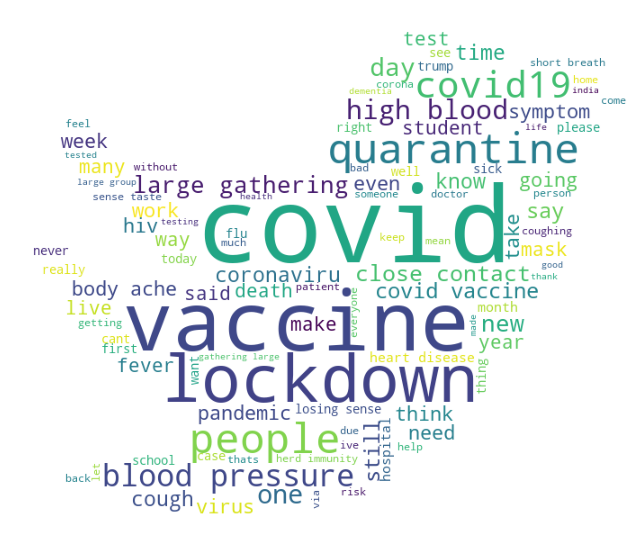}
    \caption{Word cloud representation of twitter dataset}
    \label{wordcloud}
\end{figure}

Our dataset comprises a total of 6,667 data points categorized into five classes. A Word cloud representation of the entire dataset is depicted in Fig.~\ref{wordcloud}. Notably, words such as `COVID', `vaccine', and `lockdown' prominently feature in this word cloud, signifying their prevalence within the dataset.

\begin{figure}[!htb]
      \centering
      \begin{subfigure}[b]{0.3\textwidth}
        \includegraphics[width=\textwidth]{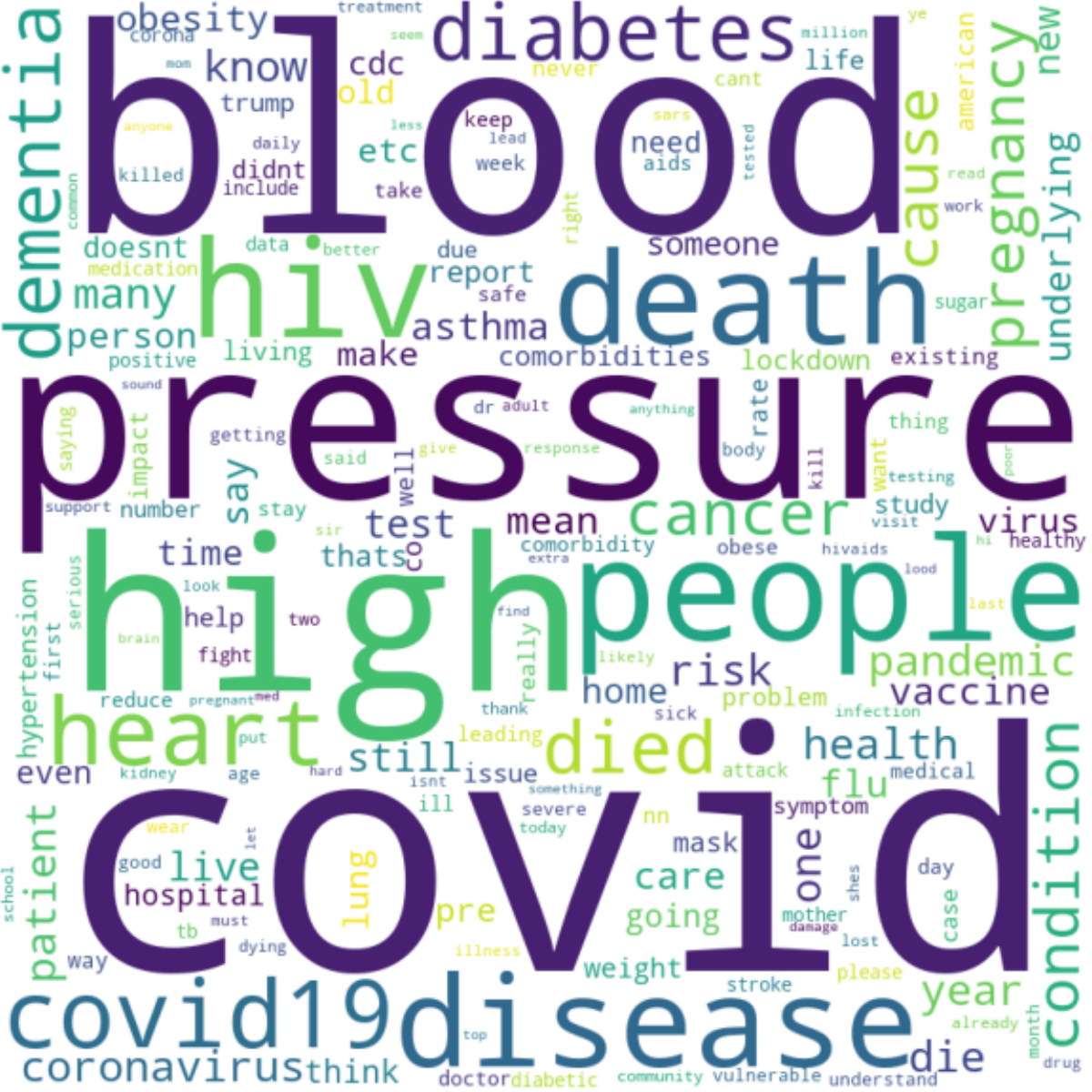}
        \caption{Health risk }
        \label{health risk}
      \end{subfigure}
      \hfill
      \begin{subfigure}[b]{0.3\textwidth}
        \includegraphics[width=\textwidth]{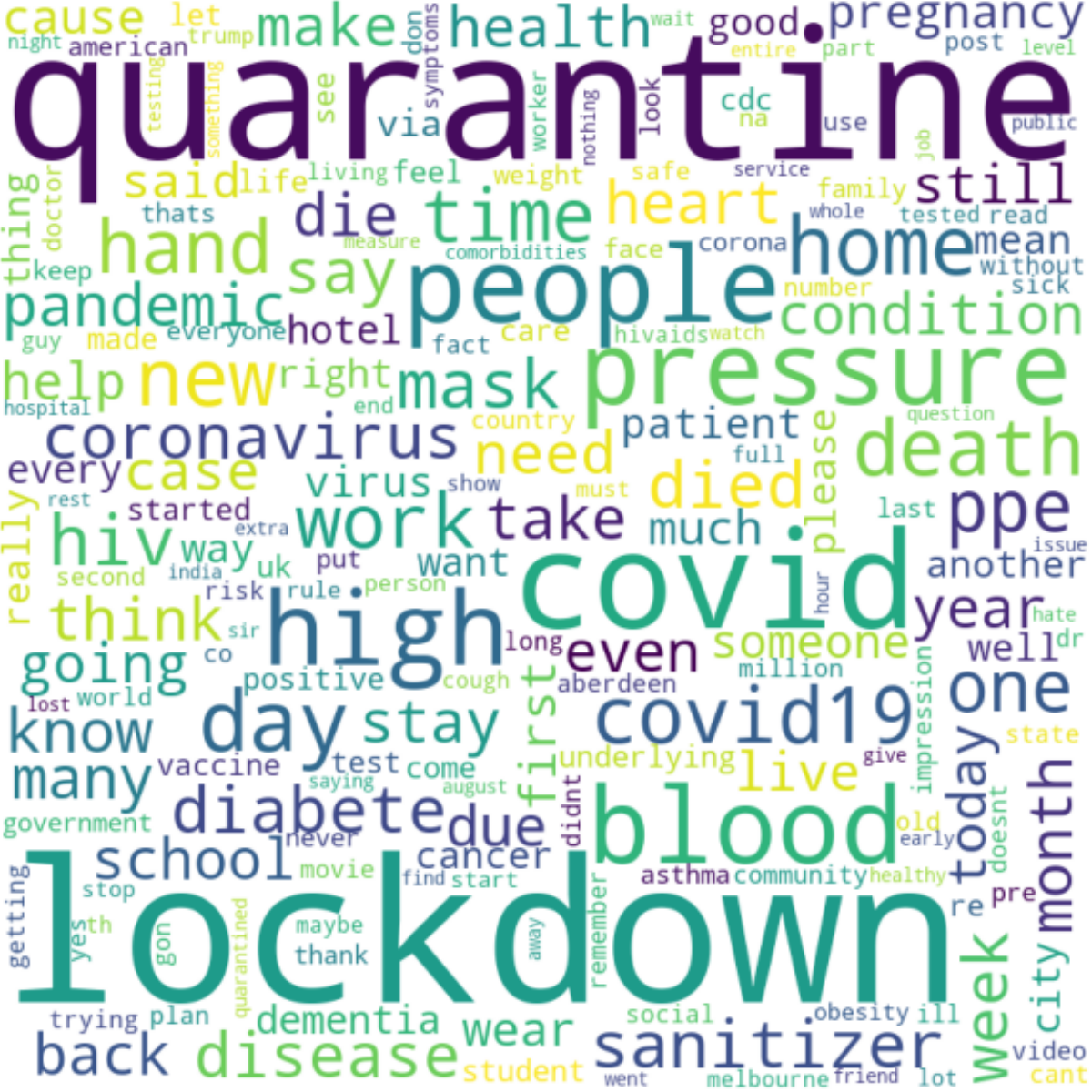}
        \caption{Prevention }
        \label{prevention}
      \end{subfigure}
    \hfill
    \begin{subfigure}[b]{0.3\textwidth}
        \includegraphics[width=\textwidth]{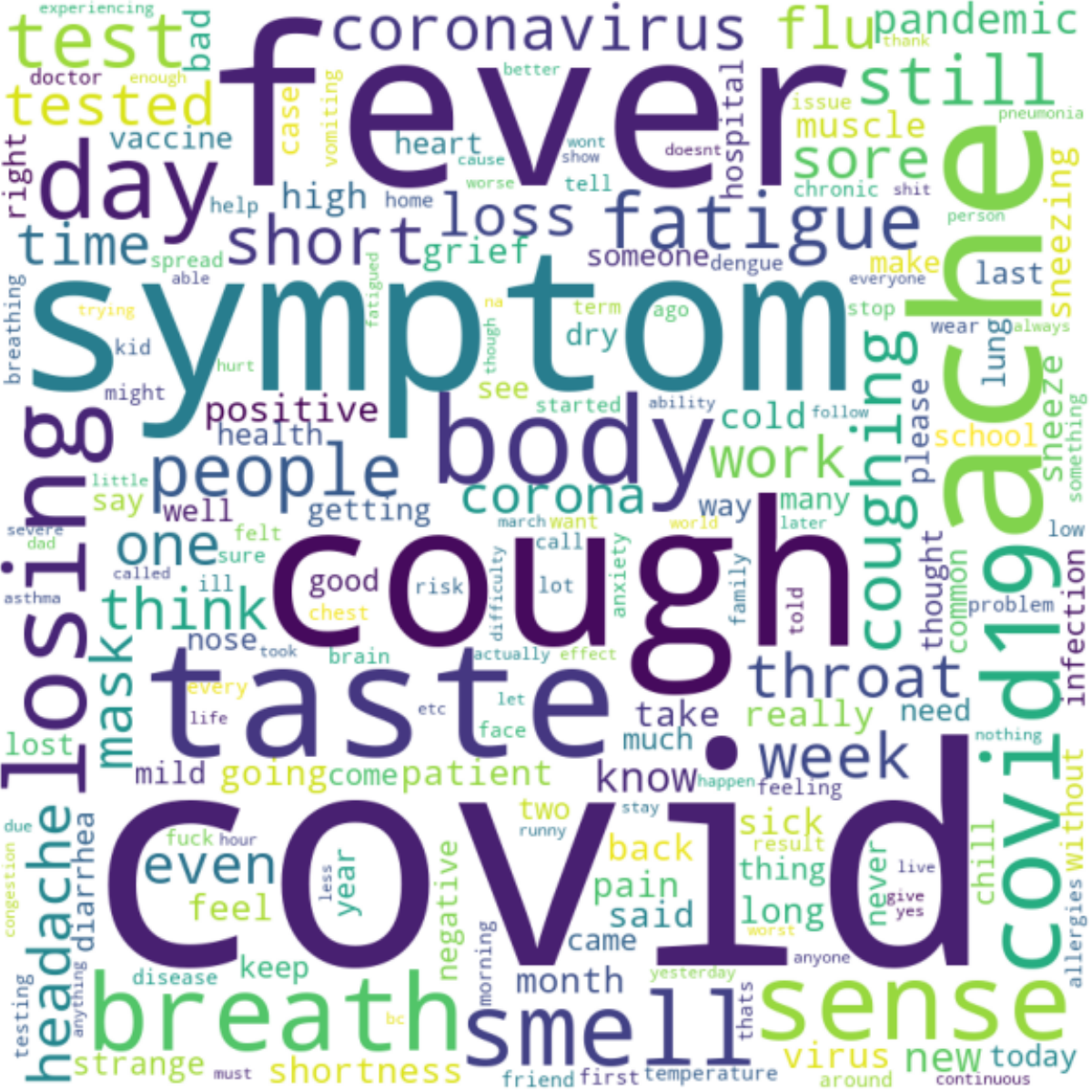}
        \caption{Symptoms }
        \label{symptoms}
      \end{subfigure}
      \begin{subfigure}[b]{0.3\textwidth}
        \includegraphics[width=\textwidth]{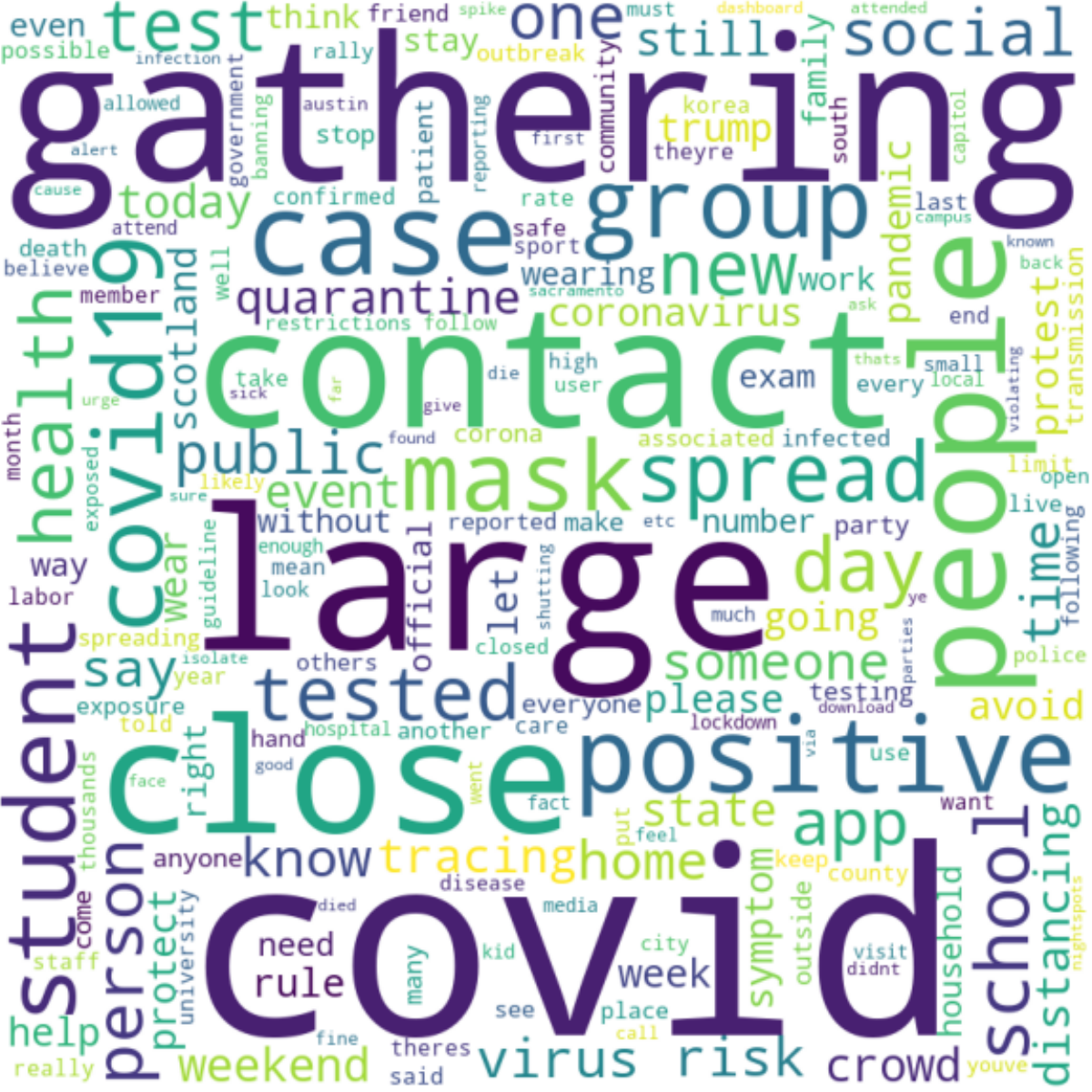}
        \caption{Transmission  }
        \label{transmission}
      \end{subfigure}
     \hspace{0.75cm}
      \begin{subfigure}[b]{0.3\textwidth}
        \includegraphics[width=\textwidth]{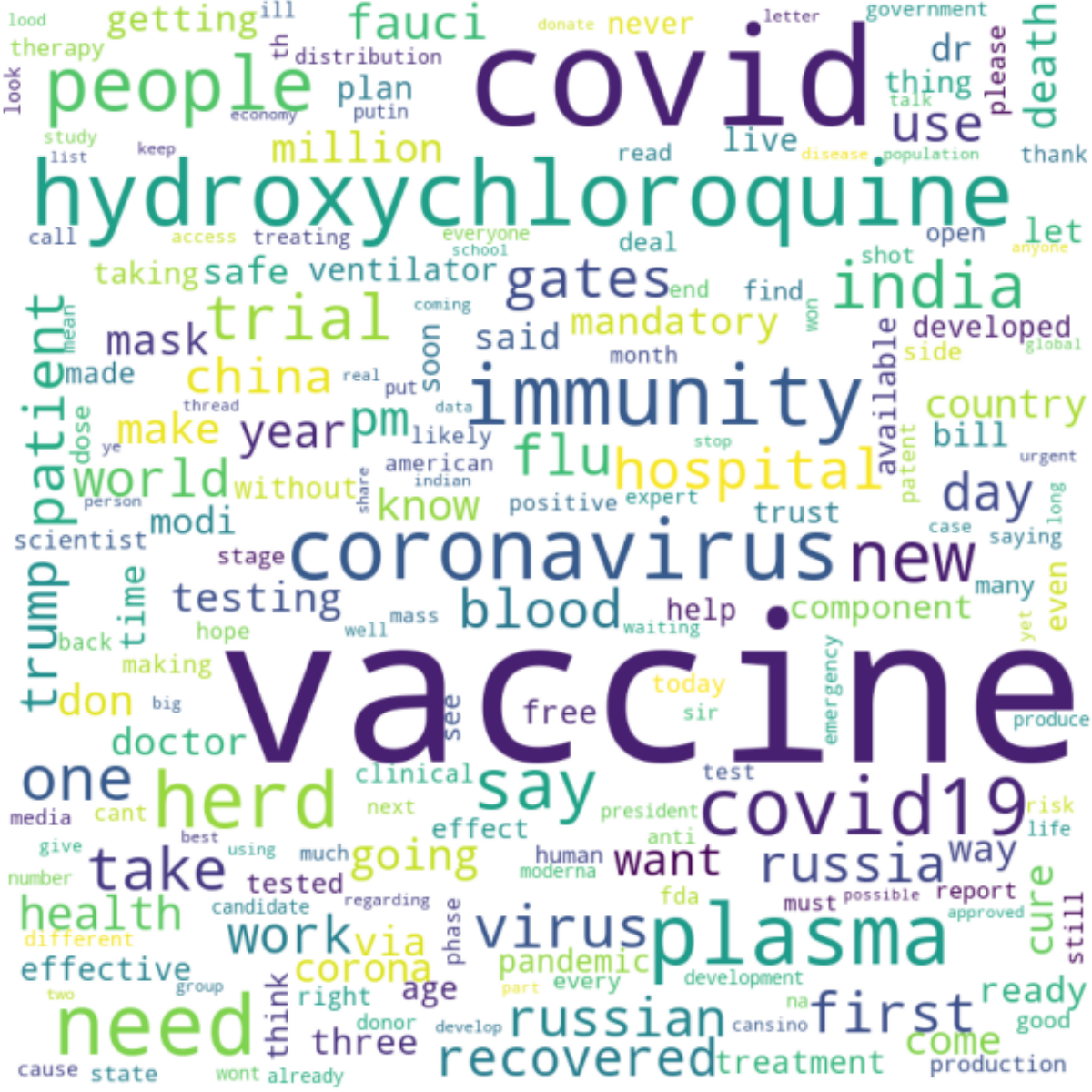}
        \caption{Treatment }
        \label{treatment}
      \end{subfigure} 
      \caption{Word cloud representation of five different classes of the COVIDHEALTH dataset}
    \end{figure}

Furthermore, individual word clouds were generated for each of the five classes, as presented in Figs.~\ref{health risk},~\ref{prevention},~\ref{symptoms},~\ref{transmission}, and ~\ref{treatment}. 
To provide a more detailed insight, we identified the top 10 most frequent words within each of these word clouds, which are summarized in Table~\ref{frequency table} for better comprehension.
Interestingly, the term `Covid' is a common occurrence across all classes. However, upon its exclusion from the \textit{health risk} class, we observed that words related to `blood pressure' and `heart disease' prominently characterize this class, aligning with the typical concerns and focus of the health risk category. In the \textit{prevention} class, terms such as `quarantine' and `lockdown' dominate the word cloud, reflecting the emphasis on preventive measures. In the \textit{symptoms} class, words like `fever', 'symptom', and `cough' exhibit higher frequency compared to other terms. Within the \textit{transmission} class, the phrase `large gathering' and the concept of `close contact' are the most prominent. Lastly, in the \textit{treatment} class, `COVID vaccine' and `vaccines' emerge as the dominant terms, underscoring the focus on treatment and vaccination within this category.
\begin{table}[!htb]
    \centering
    \caption{Top-10 frequent words from tweets in the COVIDHEALTH dataset}
    \label{frequency table}
    \resizebox{1\textwidth}{!}{
    \begin{tabular}{@{\extracolsep{2pt}}cl@{\hskip .01in}rl@{\hskip .01in}rl@{\hskip .01in}rl@{\hskip .01in}rl@{\hskip .01in}rl@{\hskip .01in}r}
 
    \hline
       \multirow{2}{*}{Rank} & 
       \multicolumn{2}{c}{Health risk} &
       \multicolumn{2}{c}{Prevention} &
       \multicolumn{2}{c}{Symptoms} &
       \multicolumn{2}{c}{Transmission} &
       \multicolumn{2}{c}{Treatment} &
       \multicolumn{2}{c}{Whole dataset}\\
    \cline{2-3}\cline{4-5}\cline{6-7}\cline{8-9}\cline{10-11}\cline{12-13}
    & Word & Count & Word & Count& Word & Count& Word & Count& Word & Count & Word & Count\\
    \hline\hline
    1 & COVID        & 508 & quarantine & 609 & Covid        & 945 & covid          & 631 & vaccine   & 927 & COVID & 2957  \\
    2 & blood        & 276 & lockdown   & 603 & fever        & 335 & large          & 428 & covid     & 480 & Vaccine        & 1001     \\
    3 & pressure     & 244 & covid      & 393 & cough        & 281 & gathering      & 357 & vaccines  & 240 & People         & 684 \\
    4 & high         & 217 & people     & 170 & aches        & 228 & close          & 267 & hydroxychloroquine & 123 & Quarantine & 656 \\
    5 & people       & 153 & pressure   & 161 & taste        & 225 & contact        & 257 & plasma    & 107 & Lockdown       & 645\\
    6 & hiv          & 137 & high       & 159 & body         & 216 & people         & 162 & coronavirus & 105 & Blood        & 507\\
    7 & heart        & 116 & like       & 104 & losing       & 188 & cases          & 83  & immunity  & 103 & High           & 444\\
    8 & disease      & 99  & home       & 91  & breath       & 162 & gatherings     & 82  & herd      & 90 & Large           & 440\\
    9 & diabetes     & 84  & get        & 82  & smell        & 152 & groups         & 77  & need      & 90 & Pressure        & 410\\
    10& dementia     & 77  & work       & 75  & fatigue      & 143 & spread         & 68  & people    & 81 & Gathering       & 363\\
    
    \hline
    
    \hline
    \end{tabular}
   }
\end{table}

\subsection{Data preprocessing}
After labelling all tweet data, we applied a series of preprocessing techniques to clean the unstructured and non-categorized dataset. This step involved the systematic application of various data preprocessing methods to refine the raw text data, including the removal of unnecessary elements such as mentions, hashtags, URLs, repeated characters, punctuation, stopwords, and text in other languages.

To begin, we utilized regular expressions to remove mentions (words preceded by the `@' symbol) and hashtags (denoted by the `\#' symbol) from all tweet data. Subsequently, we employed regular expressions to eliminate URLs present within the text.
Following this, we proceeded to remove irrelevant words, such as `rt', `brt', and newline characters (`\symbol{92}n') from the tweets. Additionally, if a word contained more than two repeated characters (e.g., `tooooo muuuuuch'), we reduced the repetitions to a maximum of two characters. While not a perfect solution due to altered spellings (e.g., `muuch'), it effectively reduces the feature space by consolidating variations like `muuuch' and `muuuuuch' into a common form, `muuch'.
Furthermore, we removed all punctuation marks, as they are typically unnecessary for text classification purposes. Stopwords, which are frequently occurring words that do not contribute unique information for classification, were also removed from the text using the Natural Language Toolkit (NLTK).
To maintain consistency, we retained only English characters in the dataset, filtering out any text in other languages. This preprocessing stage aimed to enhance the quality of the text data and prepare it for subsequent analysis and classification tasks.

\section{Machine Learning Techniques for Text Classification}\label{sec:Machine Learning Techniques for Text Classification}

\subsection{Dataset sampling techniques}\label{sec:sampling}

The dataset exhibits varying tweet counts across each class, leading to a data imbalance challenge. In such scenarios, conventional classifiers may struggle when dealing with classes of unequal sizes, favoring the larger class. To address this issue, we employed three oversampling techniques —- SMOTE (Synthetic Minority Over-sampling Technique), ADASYN (Adaptive Synthetic Sampling), and random oversampling —- as well as one under-sampling technique. This enabled us to generate a balanced dataset while retaining the original imbalanced dataset. As a result, we worked with a total of five distinct datasets.
%Number of tweets in each class are shown in Table ~\ref{tab:No of tweet in each class}

%\paragraph{SMOTE}
%A dataset is imbalanced if the classes are not approximately equally represented. 

%\paragraph{Adasyn}

%\paragraph{Random Oversampling}

%\paragraph{Under Sampling}

%\paragraph{ML model with 10 fold cross validation:}
%Cross validation is a technique to evaluate machine learning models by splitting the dataset into training set and test set for training and testing purpose respectively.
%In 10-fold cross validation process the primary dataset is randomly shuffled and then divided into 10 equal size subset.
%From these 10 subset of dataset a single subset is retained as the validation data for for testing the model performance and the remaining 9 subset are used for training purpose. This same process then repeated 10 times. In every iteration each of the subset used as the validation dataset exactly once. After the 10 iteration the folds combined to prepare a single estimation. This method is very easy to understand and it generally results in a less biased or less optimistic estimate of the model skill than other methods.

\subsection{Feature Extraction}\label{sec:featureextraction}
The number of features extracted through the feature extraction techniques is detailed in Table \ref{features extracted}, and they are described below.
\begin{table}[!htbp]
 \caption{List of extracted features with brief description}
  \centering
\resizebox{\linewidth}{!}{
\begin{tabular}{lllrl}
\hline
\multicolumn{1}{l}{Scope} &
  \multicolumn{1}{l}{\begin{tabular}[l]{@{}l@{}}Feature\\  Name\end{tabular}} &
  \multicolumn{1}{l}{Description} &
  \multicolumn{1}{r}{\begin{tabular}[r]{@{}r@{}}Feature\\  no.\end{tabular}} &
  \multicolumn{1}{l}{\begin{tabular}[l]{@{}l@{}}Output \\ Type\end{tabular}} \\ \hline \hline
Linguistic measure           & LIWC    & Measures textual features                         & 69    & Real    \\ %\hline
Word frequency               & TF-IDF  & Measures the importance of a word in a document   & 12,649 & Real    \\ %\hline
Word-category disambiguation & POS tag & Counts the number of part of speech in a document & 35    & Integer \\ \hline

\hline
\end{tabular}
}
\label{features extracted}
\end{table}
%\paragraph{Linguistic Inquiry and Word Count (LIWC):}

\textit{Linguistic Inquiry and Word Count (LIWC)}~\citep{tausczik2010psychological}, a transparent text analytics software, utilizes the original news text in the dataset to extract a comprehensive array of psychological and linguistic features. We used LIWC to extract features from 13 different dimensions: (i) summary dimension -- consists of 8 features, including word count and word per sentence, (ii) punctuation mark -- consists of 12 features, including comma, semicolon, quote, and hyphen, (iii) function words -- consists of 15 features, including pronoun, article, and conjunction, (iv) perceptual process -- consists of 4 features, including seeing, hearing, and feeling, (v) biological process -- consists of 5 features, including body, sexuality, and health, (vi) other grammar -- consists of 6 features, including interrogatives and numbers, (vii) time orientation -- consists of 3 features, such as past, present, and future, (viii) relativity -- consists of 4 features, including motion, space and time, (ix) affect -- consists of 6 features, including positive emotion, negative emotion, and anxiety, (x) personal concerns -- consists of 6 features, including achievement, leisure, and home, (xi) social -- consists of 5 features, including human, family, and friend, (xii) informal language, consists of 6 features, including filler, and swear, (xiii) cognitive process -- consists of 7 features, including certainty, insight, and inhibition.

%\paragrap{Term Frequency and Inverse Document Frequency}

\textit{Term frequency and inverse document frequency (TF-IDF)}~\citep{ahmed2017detection} weighting method is applied to the dataset to assess the importance of a term within a document. Term frequency (TF) quantifies the frequency of a term within a specific document. Conversely, the inverse document frequency (IDF) is a metric for determining the significance of a term across the entire dataset. The TF and IDF product is the weight of the TF-IDF in a given word. The higher the value of TF-IDF, the rarer it is. The TF-IDF weight is frequently used in text mining and information retrieval. 

\begin{comment}

\textbf{Term Frequency and Inverse Document Frequency (TF-IDF)}~\cite{ahmed2017detection} weighting method is used on the dataset in order to measure the significance of a term in a document. Term Frequency (TF) in a specific document is used to estimate the frequency of a term. On the other hand, the Inverse Document Frequency (IDF) is a metric for determining the importance of an item in a corpus. Let, D symbolizes N papers for the entire corpus. If ${n(t)_d}$ indicates that in document d the number of times t is shown, then TF indicated by ${TF(t)_d}$ can be calculated by equation~\ref{eqn:tf}:

\begin{equation}
\label{eqn:tf}
TF(t)_d = \dfrac{n(t)_d}{\sum_{t {'}  \epsilon d }n(t {'} )_d}
\end{equation}

And IDF, denoted by $IDF(t)_D$, can be calculated by
equation~\ref{eqn:idf}:

\begin{equation}
\label{eqn:idf}
IDF(t)_D = 1+log[N × | \{d \;  \epsilon  \; D : t \; \epsilon \; d \}|^{-1}] 
\end{equation}

The TF and IDF product is the weight of the TF-IDF in a given word. The higher the value of TF-IDF, the rarer it is. The tf-idf weight is frequently used in information retrieval and text mining. This weight is used to determine the significance of a word in a list or corpus of texts or documents. The value of a word increases in directly propotional to its intensity in the corpus, but is counterbalanced by its frequency in the corpus.
\end{comment}

%\paragraph{Part Of Speech Tagging:}

\textit{Part of speech tagging (POST)}~\citep{sharma2019combating} also known as word-category disambiguation, is used to annotate a word with a corresponding part of the speech based both on its definition and on its context for resolving lexical ambiguities. The Stanford Part of Speech Tagger was utilized for this task, and the resulting feature extracted from this process is the count of POS tags. 
Within the corpus, 33 distinct tagsets (a list of part-of-speech tags) were identified. To obtain the POS tag count, each tweet's words were tallied according to their assigned POS tags, resulting in the derivation of 33 individual features.

%\paragraph{N-gram}

% An \textit{N-gram} refers to a contiguous sequence of N characters within a larger string. Although the term can also encompass any co-occurring character combination within a string in the literature (for example, an N-gram composed of the first and third character of a word), in our context, we specifically refer to adjacent word slices. This study utilized three types of n-grams: uni-gram, bi-gram, and tri-gram. 

\subsection{Classification methods}\label{sec:classification}

Classification is a machine learning technique wherein the algorithm learns from the provided data and subsequently categorizes new observations based on this learned knowledge. This subsection covers both traditional machine learning and deep learning algorithms used in the classification process.

\subsubsection{Traditional machine learning }
Below, we will briefly discuss the traditional machine learning techniques employed in this study.
%add paragraph about what is meant by traditional ML
%Considered a subset of AI, machine learning exhibits the "learning" of human intelligence, as well as, the ability through the use of computer algorithms, to learn and analyze. These algorithms use broad sets of inputs and outputs for pattern recognition and efficient training of the system for self-referenced suggestions or decisions. The computer would be able to predict an input after sample repetition and alteration of the algorithm. The results are then compared to a set of known results in order to determine the algorithm's precision, which is then modified iteratively to perfect the ability to predict further outcomes~\citep{helm2020machine}.
    
%\paragraph{Decision Tree}
%Decision tree (DT)~\citep{safavian1991survey} uses nominal classification, which is a simple but powerful multivariate, non-parametric supervised learning process. One of the options for multi-stage decision-making is the decision tree classification, table look-up rules~\citep{haralick1976table} and table decisions to optimum decision trees~\citep{hartmann1982application} and sequential approaches. The basic concept behind any multi-stage strategy is to disassemble a complicated decision into a union of many simpler decisions, in the hope that the finished solution will be the desired one.

Decision tree (DT)~\citep{safavian1991survey} is a powerful machine learning algorithm that is widely used for both classification and regression tasks. It is a graphical representation of a decision-making process that resembles a tree structure with nodes and branches. Each node represents a feature or attribute, and each branch represents a decision or outcome based on that feature. The decision tree works by recursively splitting the data into subsets based on the most informative features at each node, effectively partitioning the data into categories or predicting numerical values. This process continues until a stopping criterion is met, such as a predefined depth or a minimum number of data points in a node.
Decision trees are known for their interpretability and ease of understanding. They are used in a wide range of applications, from finance to healthcare, and are often a fundamental building block in more complex machine learning models.

Adaboost~\citep{freund1997decision}. Boosting is a machine-learning technique based on a combination of several relatively weak and inexact rules for constructing a highly accurate Prediction Law. AdaBoost, unlike boost-by-majority, combines the weak hypotheses by summing their probabilistic predictions. In a real-valued neural network summing the outcomes of the networks and then selecting the best prediction performs better than selecting the best prediction of each network and then combining them with a majority rule~\citep{drucker1993boosting}.

Random forest (RFs)~\citep{Freund96experimentswith} are a collection of tree predictors in which the values of a random vector sampled independently and with the same distribution for all trees in the forest are used to predict the behavior of each tree. If the number of trees in a forest grows larger, the generalization error converges to a limit. The intensity of individual trees in the forest and the correlation between them determine the generalization error of a forest of tree classifiers. When a random set of features is used to separate each node, the error rates are comparable to AdaBoost, which theoretically reduces any learning algorithm error that consistently generates classifiers whose performance is a little better than random assumption but more robust in terms of noise. Internal estimates are used to track error, power, and correlation~\citep{breiman2001random}.

Stochastic gradient descent (SGD)~\citep{bottou201113} is an iterative approach for optimizing an objective function with sufficient smoothness properties (e.g., differentiable or sub-differentiable). It replaces the actual gradient which is calculated from the entire data set by an estimated gradient which is calculated from a randomly selected subset of the data. So it can be regarded as a stochastic approximation of gradient descent optimization. This reduces the computational burden and achieves quicker iterations in trade for a lower convergence rate, particularly in the case of high-dimensional optimization problems.
        
K-nearest neighbour~\citep{cunningham2020k}. The intuition behind the k-nearest neighbor (kNN) classification is very simple: examples are categorized by their closest neighbor's class. More than one neighbor must also be taken into account so that the method is more generally called kNN classification, where k closest neighbors are used in the class determination. Because the training examples are required during run time, i.e., they must be in memory during run time, often they are referred to as memory-based classification. Since the induction is delayed, a lazy learning technique is considered.

Logistic regression~\citep{menard2002applied} is a supervised classification algorithm used to estimate a target variable's probability. The type of objective or dependent variable is dichotomous, meaning that only two classes are possible. The dependent variable, in simple words, is binary in nature with either $1$ (this means success/yes) or $0$ (this indicates failure/no). A logistic regression model predicts $P (Y=1)$ as a function of $X$ mathematically.

Linear SVC~\citep{perez2001svc}. The aim of the Linear SVC is to fit the data that is provided, returning a hyperplane that is ``best fit" and divides or classifies data. From there, some features can be fed to the classifier after receiving the hyperplane to see what the predicted class is. This makes this particular algorithm more acceptable for use, although this can be used for many situations.
    
\subsubsection{Deep learning}

Deep learning makes it possible to learn data representation with multiple abstraction levels through computational models that consist of various processing layers. This increases the state-of-the-art in speech identification, the recognition of visual artifacts, the detection of objects, and many other domains including drug discoveries and genomics dramatically. Deep learning detects complex structures in large data sets with the use of the backpropagation algorithm to tell how the computer changes its membership functions, which are used to calculate the representation in each layer from the representation in the previous layer~\citep{lecun2015deep}.

%\paragraph{CNN} 
%In a neural network, where neurons are fed inputs which then neurons consider the weighted sum over them and pass it by an activation function and passes out the output to next neuron.

%Now, a convolutional neural network is different from that of a neural network because it operates over a volume of inputs. Each layer tries to find a pattern or useful information of the data.

%It will be different depending on the task and data-set we work on. There are some terms in the architecture of a convolutional neural networks that we need to understand before proceeding with our task of text classification.

%Convolution: It is a mathematical combination of two relationships to produce a third relationship. Joins two sets of information.

%Convolution over input:Input data is slided over the convolution to extract features by applying a filter/ kernel. This is important in feature extraction. There are some parameters associated with that sliding filter like how much input to take at once and by what extent should input be overlapped.

\textit{Convolutional Neural Networks (CNNs)}~\citep{o2015introduction} are similar to conventional artificial neural networks (ANNs) since they consist of self-optimizing neurons~\citep{abiodun2018state}. Each neuron still receives input and carries out an action (such as a scalar product followed by a non-linear function) –- the basis for countless ANNs. The entire network will still express a single perceptive score feature from input raw text through to the final output of the class score (the weight). The final layer contains class-related loss functions. CNNs are comprised of three types of layers. These are convolutional layers, pooling layers, and fully-connected layers. When these layers are stacked, a basic CNN architecture has been formed. By calculating the scalar product between its weights and the region linked to the input volume, the convolution layer determines the output of neurons that are connected to local regions of the input. The pooling layer will then simply perform downsampling along the spatial dimensionality of the given input, further reducing the number of parameters within that activation. The fully connected layers try to generate class scores from the activations for classification purposes. 

We have used an embedding dimension of size 100 in our CNN architecture. The optimizer used in our CNN is RMSProp. Loss accuracy is measured by binary cross-entropy. We have used one dimensional convolution layer for this work. We add two layers to our CNN architecture. We have used relu activation in the first layer of the architecture and sigmoid activation in the second layer. Sixty epochs have been considered for the training dataset.

\textit{Recurrent Neural Network (RNN)}~\citep{sherstinsky2020fundamentals} is a feedforward neural network with an internal memory that is a generalization of the feedforward neural network. RNN is recurrent in nature since it executes the same function for each data input, and the current input's outcome is dependent on the previous computation. The output is replicated and transmitted back into the recurrent network when it is created. It evaluates the current input as well as the output it has learned from the prior input when making a decision.

In the architecture of RNN, we have used 0.3 dropouts between the nodes of the convolution layer. We have used rmsprop optimizer for layers in RNN architecture. We have used sigmoid activation for the RNN layer. We have used an embedding dimension of size 100 for RNN architecture.

\textit{Long Short-Term Memory Networks (LSTMs)}~\citep{hochreiter1997long} represent a type of Recurrent Neural Network (RNN) designed to capture long-term dependencies. LSTMs gained widespread popularity and have proven effective in various applications. Unlike ordinary RNNs composed of simple repeating modules, LSTMs utilize a more complex structure, particularly in their cell state, depicted as a horizontal line traversing the diagram.
The cell state in LSTMs functions akin to a conveyor belt, allowing data to flow unchanged with minimal linear interactions. Notably, LSTMs regulate the information flow through structures known as gates. Gates act as a selective mechanism, enabling the controlled addition or deletion of information from the cell state. Comprising a sigmoid neural net layer and a point-wise multiplication operation, gates play a crucial role in determining which information is allowed to pass through.
To update the cell state, LSTMs employ an input gate layer (sigmoid) that selects values for updating and a tanh layer generating a vector of new candidate values~\citep{jelodar2020deep}. These components are combined in the subsequent phase to create a comprehensive state update.

In this study, a sequential LSTM model is utilized. The architecture employs a softmax activation function and the Adam optimizer. Categorical cross-entropy is used to measure loss accuracy. The LSTM layer incorporates a dropout rate of 0.2, and recurrent dropout is also set at 0.2.

The \textit{Bidirectional Encoder Representations from Transformers (BERT)}~\citep{devlin2018bert} model is structured in two distinct stages: pre-training and fine-tuning. During pre-training, the model is trained on a wide, unlabeled corpus. Subsequently, in the fine-tuning phase, all parameters are further adjusted using labeled data for specific tasks, building upon the knowledge gained during pre-training. The initial parameters for fine-tuning are derived from the pre-trained model.
The BERT architecture is rooted in a bidirectional transformer multi-layer encoder design, which eliminates the need for recurrence and instead leverages a mechanism based on attention for establishing global dependencies between input and output~\citep{vaswani2017attention}.

Two primary types of pre-training are employed: Masked Language Modeling (MLM) and Next Sentence Prediction (NSP). In MLM, a portion of input tokens are randomly masked, and the model learns to predict these masked tokens, thus fostering deep bidirectional representations. The NSP task can be generated from any monolingual corpus, facilitating the training process. BERT models can then be fine-tuned for various downstream tasks using the transformer's self-attention mechanism~\cite{devlin2018bert}.

In this study, the BERT model is trained using a batch size of 32 for training and 8 for evaluation. The learning rate utilized in the architecture is set to $1e-5$. A warm-up proportion of 0.5 is applied. The maximum sequence length is limited to 50. The BERT model is trained for a total of 3 epochs.

\subsection{Evaluation metrics}
The following assessment measures were used to assess the classification performance: accuracy, precision, recall, and F1 score. The definitions of accuracy, precision, recall, and F1 score are as follows:
    
\[ Accuracy = \dfrac{TP + TN}{TP+TN+FP+FN}\]
\[ Precesion = \dfrac{TP }{TP + FP} \]
\[ Precesion = \dfrac{TP }{TP + FN} \]
\[ F1 = 2 \times \dfrac{Precesion \times Recall}{Precesion + Recall} \]
Here TP, TN, FP, and FN stand for true positive, true negative, false positive, and false negative respectively.

\section{Experiments and Results}\label{sec:Experiments and Results}
In this section, we will present the findings of our study, focusing on dataset settings and the results of both traditional machine learning and deep learning algorithms.
In this study, we employed seven different traditional machine learning algorithms: Decision Tree, Random Forest, Stochastic Gradient Descent, K-nearest neighbor, Adaboost, Logistic Regression, and Linear SVC. For deep learning algorithms, we exclusively utilized the TF-IDF feature extraction method, excluding the LIWC and POS tag feature extraction methods. This exclusion was due to the poor performance of LIWC and POS tag features in the context of deep learning, resulting in significantly lower accuracy scores. The TF-IDF method yielded 12,649 features, while LIWC and POS tag methods produced only 69 and 35 features, respectively. When compared to the vast number of features from TF-IDF, the features extracted by LIWC and POS tag methods were minimal. Consequently, these two feature extraction methods did not yield superior results; instead, they introduced additional computational costs without significant benefits.

\subsection{Dataset Settings for machine learning algorithm}
%\subsubsection{Dataset Settings}
Table~\ref{tab:No of tweet in each class} provides an overview of our original dataset. In addition to this original imbalanced dataset, we created four additional datasets using the oversampling and undersampling techniques described in Subsection~\ref{sec:sampling}. Our goal was to explore whether these techniques could enhance accuracy.

For the traditional machine learning approach, we initially divided the original dataset into training and testing subsets. In this split, 20\% of the data was allocated for testing, while the remaining 80\% was used for training. Importantly, the sampling techniques were exclusively applied to the training data.

The initial pre-processed dataset comprised a total of 6,667 data points, with an inherent class imbalance. This data was partitioned into 80\% for training and 20\% for testing. Through various oversampling techniques, we augmented the dataset to a total of 9,544 data points, which were subsequently split into 80\% training and 20\% testing subsets. To address the class imbalance, we generated an under-sampled dataset and two oversampled datasets—employing random oversampling, ADASYN, and SMOTE techniques—based on the original data. Additionally, we crafted another dataset by applying undersampling to the original dataset, resulting in 640 data points in each class for training and 160 data points in each class for testing, as detailed in Table~\ref{dataset setting for traditional ml}.

\begin{table}[!htbp]
\caption{Dataset setting for traditional machine learning algorithm}
\centering
\begin{tabular}{lrrrrrr}
\hline
\multirow{2}{*}{Dataset} & \multicolumn{5}{c}{Class}                                     & \multirow{2}{*}{Total} \\ \cline{2-6}

                         & Symptoms & Treatment & Health risk & Transmission & Prevention &                        \\ \hline\hline
Original               & 1,402     & 1,439      & 978         & 802          & 2,046       & 6,667                   \\ 
Oversampling             & 2,046     & 2,046      & 2,046        & 2,046         & 2,046       & 10,230                   \\ 
Undersampling            & 800      & 800       & 800         & 800          & 800        & 4,000                   \\ \hline

\hline
\end{tabular}

\label{dataset setting for traditional ml}
\end{table}

\subsection{Results for machine learning algorithm}
%In this sub section we will present the result of the traditional machine learning approach with their precision score, recall score, F-measure and accuracy. In this study the five class classification problem determines whether the tweet is related to prevention, treatment, symptoms, health risk or transmission category.

%In Table~\ref{cross_validation_precision} the classification accuracy, precision, recall, and F1 score are presented. We have achieved the maximum accuracy of 86.28\% for the random oversampling dataset using the stochastic gradient descent algorithm. We have achieved the highest precision score for stochastic gradient descent (86.55\%), the highest recall for stochastic gradient descent (87.28\%), and the highest F1 score for stochastic gradient descent (86.34\%). The lowest accuracy is achieved by using an under-sampling dataset in all of the algorithms because of the lower amount of sample data. In the case of other datasets except under-sampling every dataset performed moderately on the algorithms. 10-fold cross-validation has been used to evaluate all algorithms.

Table~\ref{cross_validation_precision} presents the classification accuracy, precision, recall, and F1 score, from 10-fold cross-validation using traditional machine learning algorithms. The maximum accuracy, at 86.28\%, was achieved using the stochastic gradient descent algorithm on the random oversampling dataset. Likewise, the highest precision score, 86.55\%, was attained with stochastic gradient descent. The highest recall score of 87.28\% was also achieved with stochastic gradient descent, and the highest F1 score, 86.34\%, was likewise obtained with this algorithm.
Conversely, the lowest accuracy scores were observed when using the under-sampling dataset, across all algorithms. This is due to the smaller amount of sample data available in this particular dataset. However, for the other datasets, excluding under-sampling, every dataset exhibited moderate performance 

%new table for 10fold

% \usepackage{multirow}

% \usepackage{multirow}
\begin{table}[!htbp]
\caption{Precision, recall, F1-score and accuracy score of different traditional machine learning algorithm (10 fold cross validation)}
\centering
\resizebox{.64\linewidth}{!}{
\begin{tabular}{llcccc}
\hline
                 Dataset       & ML Technique                & Precision & Recall & F1           & Accuracy \\ \hline\hline
\multirow{7}{*}{\begin{tabular}[c]{@{}l@{}}Original  \end{tabular}}      & Decision Tree & 83.15 & 82.57 & 82.71 & 82.88 \\ 
                        & Random Forest               & 84.58     & 83.47  & 83.95          & 83.86    \\ 
                        & Stochastic Gradient Descent & 84.79     & 84.79  & 84.72          & 84.90    \\ 
                        & K-nearest Neighbour         & 74.91     & 73.18  & 73.81          & 73.96    \\ 
                        & Adaboost                    & 82.38     & 78.05  & 77.42          & 77.57    \\ 
                        & Logistic Regression         & 84.64     & 84.10  & 54.25          & 54.32    \\ 
                        & Linear SVC                  & 85.56     & 86.44  & 85.89          & 85.94    \\ \hline
\multirow{7}{*}{SMOTE }  & Decision Tree              & 82.54     & 82.54  & 82.47          & 82.50    \\  
                        & Random Forest               & 84.38     & 83.56  & 83.89          & 83.94    \\  
                        & Stochastic Gradient Descent & 84.49     & 86.10  & 84.95          & 84.98    \\ 
                        & K-nearest Neighbour         & 67.96     & 69.48  & 57.73          & 57.54    \\  
                        & Adaboost                    & 82.90     & 79.54  & 77.58          & 77.74    \\ 
                        & Logistic Regression         & 84.83     & 84.55  & 84.92          & 84.94    \\ 
                        & Linear SVC                  & 85.71     & 86.94  & \textbf{86.13} & 86.12    \\ \hline
\multirow{7}{*}{ADASYN} & Decision Tree               & 82.59     & 82.16  & 82.27          & 82.33    \\  
                        & Random Forest               & 83.90     & 82.97  & 83.36          & 83.32    \\ 
                        & Stochastic Gradient Descent & 81.66     & 83.26  & 82.98          & 82.90    \\ 
                        & K-nearest Neighbour         & 64.88     & 25.91  & 15.40          & 15.46    \\ 
                        & Adaboost                    & 81.40     & 77.32  & 76.64          & 76.72    \\ 
                        & Logistic Regression         & 82.06     & 83.90  & 83.38          & 83.50    \\ 
                        & Linear SVC                  & 85.33     & 86.40  & \textbf{86.13} & \textbf{86.23}    \\ \hline
\multirow{7}{*}{\begin{tabular}[c]{@{}l@{}}Random\\ Over\\ Sampling\end{tabular}} & Decision Tree & 82.57 & 82.55 & 82.47 & 82.54 \\ 
                        & Random Forest               & 84.66     & 84.65  & 84.47          & 84.54    \\  
                        & Stochastic Gradient Descent & 84.64     & 86.26  & 85.04          & 85.50    \\ 
                        & K-nearest Neighbour         & 70.65     & 72.23  & 70.32          & 70.35    \\  
                        & Adaboost                    & 81.60     & 75.67  & 74.88          & 74.95    \\ 
                        & Logistic Regression         & 84.64     & 86.11  & 85.03          & 85.08    \\ 
                        & Linear SVC                  & 85.26     & 86.50  & 85.69          & 85.75    \\ \hline
\multirow{7}{*}{\begin{tabular}[c]{@{}l@{}}Under\\ Sampling\end{tabular}}         & Decision Tree & 41.00    & 42.00    & 43.00    & 43.20 \\  
                        & Random Forest               & 56.00        & 57.00     & 56.00             & 56.15    \\  
                        & Stochastic Gradient Descent & 57.00        & 58.00     & 58.00             & 58.02    \\ 
                        & K-nearest Neighbour         & 46.00        & 50.00     & 50.00             & 50.07    \\ 
                        & Adaboost                    & 52.00        & 53.00     & 54.00             & 54.29    \\  
                        & Logistic Regression         & 50.00        & 51.00     & 51.00             & 51.20    \\ 
                        & Linear SVC                  & 45.00        & 47.00     & 47.00             & 47.14    \\ \hline
                        
                        \hline
\end{tabular}
}
\label{cross_validation_precision}
\end{table}

\subsection{Dataset settings for deep learning algorithm}

%\subsubsection{Dataset settings}

For deep learning algorithms, we've prepared two types of datasets: a balanced dataset (not augmented) and a balanced dataset (augmented), both derived from the original imbalanced dataset. The dataset was divided into training (70\%), validation (20\%), and testing (10\%) subsets.
In the balanced (not augmented) dataset, the training dataset includes 640 samples for each class, while the validation and testing datasets consist of 80 samples for each class. On the other hand, the balanced (augmented) dataset features 1,600 samples for each class in the training set, with 200 samples for each class in both the validation and testing sets. Detailed dataset settings for deep learning algorithms can be found in Table~\ref{dataset setting for deep learning ml}.

%Table ~\ref{tab:tweet partition} describes...
\begin{table}[!htbp]
\caption{Dataset setting for deep learning algorithm}
\centering
\resizebox{1\textwidth}{!}{
\begin{tabular}{llrrrrrr}
\hline
\multicolumn{2}{c}{\multirow{2}{*}{Dataset}}                                          & \multicolumn{5}{c}{Class}               & \multirow{2}{*}{Total} \\ \cline{3-7}
\multicolumn{2}{c}{}                      & Symptoms & Treatment & Health risk & Transmission & Prevention &  \\ \hline\hline

\multirow{3}{*}{Original}                 & Training (70\%) & 971 & 1009 & 690 & 554 & 1442 & 4666 \\  
& Validation (20\%) & 293 & 285 & 191         & 167 & 397 & 1333 \\ 
& Testing (10\%) & 138 & 145 & 97 & 81 & 206 & 667 \\ \hline

\multirow{3}{*}{\begin{tabular}[l]{@{}l@{}}Balanced\\ (not augmented)\end{tabular}} & Training & 640 & 640 & 640 & 640 & 640 & 3200 \\ 
& Validation & 80 & 80 & 80 & 80 & 80 & 400 \\ 
& Testing & 80 & 80 & 80 & 80 & 80 & 400 \\ \hline
\multirow{3}{*}{\begin{tabular}[l]{@{}l@{}}
Balanced\\ (augmented)\end{tabular}} & Training & 1600 & 1600 & 1600 & 1600 & 1600 & 8000 \\ 
& Validation & 200 & 200 & 200 & 200 & 200 & 1000 \\  
& Testing & 200 & 200 & 200 & 200 & 200 & 1000 \\ 
\hline

\hline

\end{tabular}
}
\label{dataset setting for deep learning ml}
\end{table}

\subsection{Results for deep learning algorithm}
Four deep learning algorithms, namely LSTM, CNN, RNN, and BERT, were employed for each dataset. The results for the deep learning algorithms are reported in Table~\ref{deep learning}.
In the case of the original dataset, the BERT algorithm yielded the most promising results, achieving a training accuracy of 89\%, a validation accuracy of 74\%, and an F1 score of 82\% for the testing dataset. Although LSTM showed the highest testing accuracy, it exhibited comparatively lower validation accuracy, indicating overfitting due to the limited dataset size. CNN and RNN produced F1 scores of 73\% and 77\%, respectively, in the testing dataset.
For the balanced (not augmented) dataset, the BERT algorithm produced the highest F1 score, reaching 88\%. CNN and RNN achieved F1 scores of 83\% and 79\%, respectively. However, the substantial gap between training and validation accuracy for LSTM implied overfitting.
In the balanced (augmented) dataset, BERT and RNN achieved F1 scores of 89\% and 87\%, respectively, while CNN delivered the highest F1 score of 90\%. LSTM exhibited an F1 score of 89\% in the testing dataset, but its validation accuracy was notably lower at 19.60\%.

\begin{table}[!htbp]
\caption{Result table for testing dataset of deep learning algorithms}
\centering
\begin{tabular}{llcccc}
\hline
Dataset                   & Model name & Precision & Recall & F1             & Accuracy \\ \hline \hline
\multirow{4}{*}{Original} & LSTM       & 89.32     & 89.86  & 89.26          & 89.20     \\ 
                          & CNN        & 73.36     & 73.24  & 73.38          & 73.29    \\ 
                          & RNN        & 77.40     & 77.40  & 77.33          & 77.25    \\ 
                          & BERT       & 83.40     & 82.16  & 82.38          & 82.29    \\ 
                          \hline
\multirow{4}{*}{\begin{tabular}[c]{@{}l@{}}Balanced\\ (not augmented)\end{tabular}} & LSTM & 78.96 & 79.34 & 79.23 & 79.28 \\ 
                          & CNN        & 83.34     & 83.32  & 83.26          & 83.20    \\  
                          & RNN        & 79.39     & 79.38  & 79.41          & 79.36    \\ 
                          & BERT       & 87.83     & 88.54  & 88.20          & 88.09    \\ 
                          \hline
\multirow{4}{*}{\begin{tabular}[c]{@{}l@{}}Balanced\\ (augmented)\end{tabular}}     & LSTM & 89.47 & 89.40 & 89.37 & 89.40 \\ 
                          & CNN        & 90.48     & 90.50  & \textbf{90.43} & 90.50    \\  
                          & RNN        & 87.42     & 87.60  & 87.46          & 87.60    \\ 
                          & BERT       & 88.62     & 90.63  & 89.16          & 89.36    \\ 
                          \hline
                          
                          \hline
\end{tabular}
\label{deep learning}
\end{table}

\subsection{Best result}
In this subsection, we present the most notable results achieved by both traditional machine learning and deep learning algorithms.

\begin{figure}[!htbp]
    \centering
    \includegraphics[width=9cm]{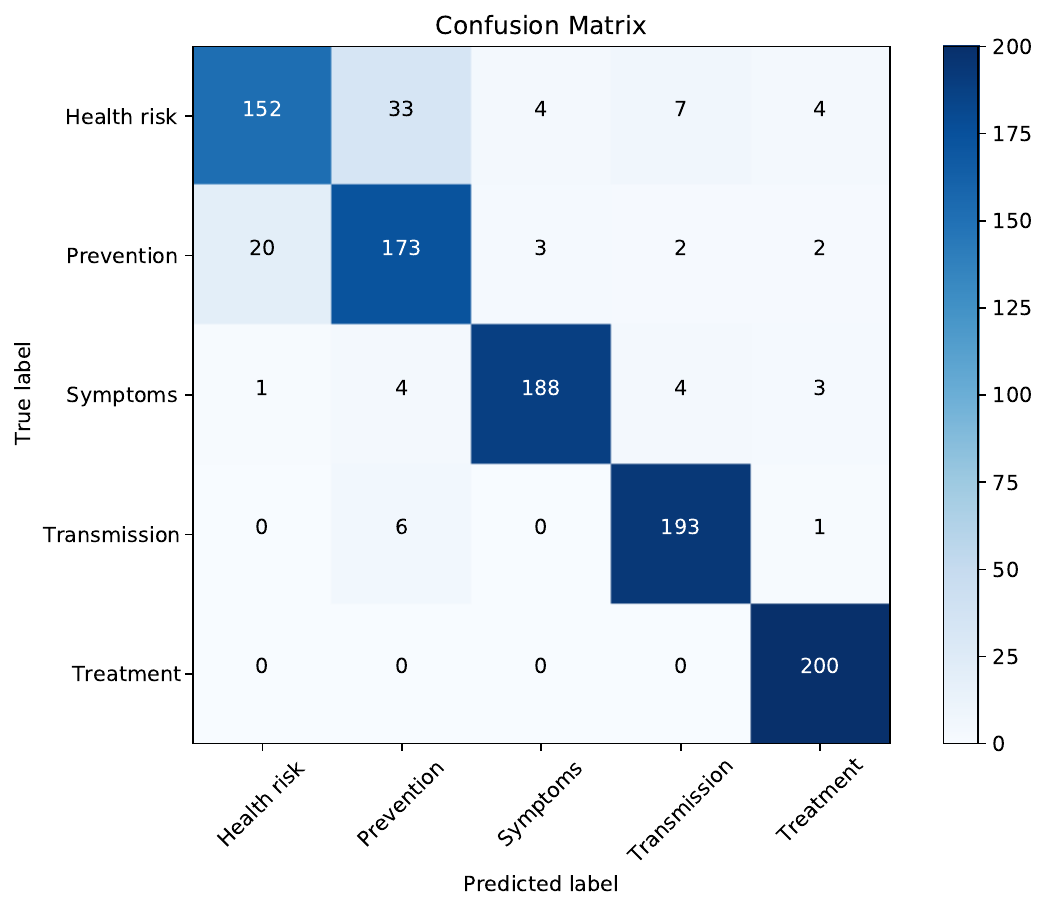}
    \caption{Confusion matrix for Linear SVC algorithm}
    \label{confusion matrix}
\end{figure}

Within the realm of traditional machine learning algorithms, the stochastic gradient descent algorithm delivered the best results. The corresponding confusion matrix is displayed in Figure~\ref{confusion matrix}, with an F1 score of 86.34\%. Precision, recall, and accuracy for this algorithm reached 86.55\%, 87.28\%, and 86.28\%, respectively. Stochastic gradient descent operates by calculating the cost of a single data point and its corresponding gradient, updating weights incrementally. This approach eliminates the need to assess all training examples simultaneously, making it highly efficient and suitable for accommodating the dataset, which contributes to its strong performance.

\begin{figure}[!htb]
    \centering
    \includegraphics[width=9cm]{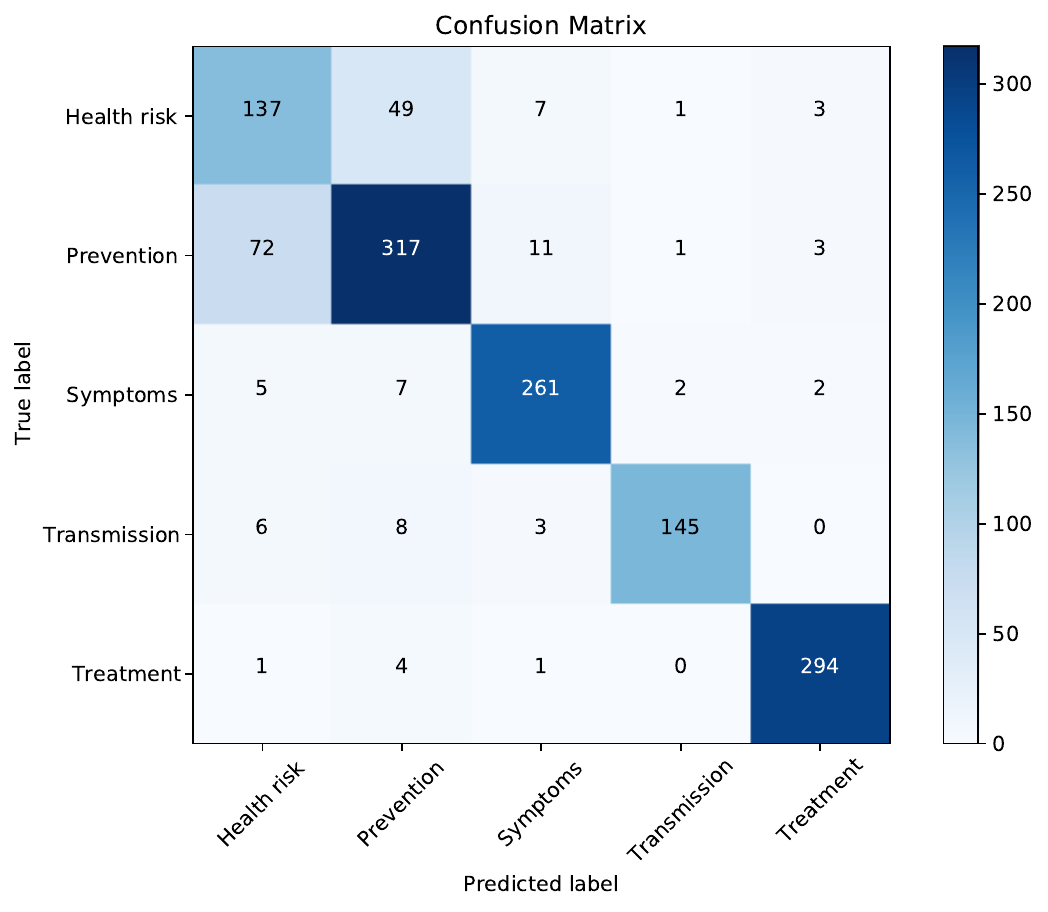}
    \caption{Confusion matrix for CNN model}
    \label{confusion matrix cnn}
\end{figure}

Among the deep learning algorithms, the CNN model performed exceptionally well, yielding the best results when applied to the balanced augmented dataset. Training accuracy reached 88.20\%, while validation accuracy stood at 89\%. Testing accuracy also demonstrated a remarkable 89\%. 
%The training loss for CNN was 0.1122, and the validation loss was 0.4031. The optimal results were attained after training the model for 60 epochs. 
The corresponding confusion matrix is presented in Figure~\ref{confusion matrix cnn}, with an impressive F1 score of 90\%. The macro average for precision, recall, and accuracy were 90\%, 90\%, and 90.6\%, respectively. CNN excels in tasks that demand feature identification in text, such as identifying emotional expressions, abusive language, and named entities. Sentiment analysis, spam detection, and topic categorization are some of its prime applications. CNNs are highly effective in extracting local, position-invariant features, making them a logical choice for sentiment classification and other text-based classification tasks.

%In this subsection different accuracy for different dataset are presented. Table ~[\ref{tab:F1-score of different traditional ML algorithm on different dataset}] describes different accuracy for different datasets for traditional machine learning algorithm

%Fig~\ref{fig:LinearSVC} and Fig~\ref{fig:LogisticRegression} shows how the F1 score is increasing with the increment of no of features. Here one gram, bigram and trigram method is also observed and we achived maximum score for bigram (Fig~\ref{fig:LinearSVC} Linear SVC) and for (Fig~\ref{fig:LogisticRegression} Logistic Regresssion) max F1 score found in trigram.

%%%   N gram graph 

 % \begin{figure}[!htb]
 %      \centering
 %      \begin{minipage}[b]{0.45\textwidth}
 %        \includegraphics[width=\textwidth]{image/accuracy curve.png}
 %        \caption{accuracy curve}
 %        \label{accuracy curve}
 %      \end{minipage}
 %    \hfill
 %      \begin{minipage}[b]{0.45\textwidth}
 %        \includegraphics[width=\textwidth]{image/loss curve.png}
 %        \caption{loss curve}
 %        \label{loss curve}
 %      \end{minipage}
 %    \end{figure}

%In other studies, researchers only make COVID-19 related twitter dataset for further scientific research~\citep{banda2020large} like detect misinformation from twitter during COVID-19~\citep{shahi2021exploratory}. Researchers have labeled tweet dataset into six classes which are mark, group labeling, responsibility, peril, misinformation, conspiracy theory in paper~\citep{li2020constructing}.

\section{Web Application}
\label{sec:application}

We have leveraged our top-performing CNN model to create a user-friendly web application\footnote{\url{https://github.com/Bishal16/COVID19-Health-Related-Data-Classification-Website}}, which can greatly assist individuals in classifying various COVID-19 related text data into the aforementioned five distinct categories. This web application employs the CNN model we developed as its backend processing engine. Users can input text, which the application then sends to the model for analysis and classification.

Using the application is straightforward: users need to navigate to the website, input or paste the text they wish to classify, and click the ``Predict" button to receive the output. To enhance user convenience, we have also developed a Google Chrome browser extension\footnote{\url{https://github.com/Bishal16/Google-Chrome-Extension_Covid19-Health-Related-Data-Classifier}}, making it even easier to access the website. Upon clicking the extension, a popup labeled "Go for classification" will appear, and clicking this button opens a new tab with the classification website.

\begin{figure}[!htbp]
      \centering
      \begin{subfigure}[b]{0.4\textwidth}
        \includegraphics[width=\textwidth]{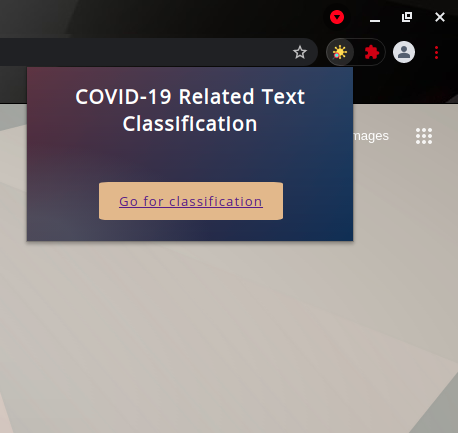}
        \caption{Google Chrome extension }
        \label{extension}
      \end{subfigure}
      \hspace{1cm}
      \begin{subfigure}[b]{0.4\textwidth}
        \includegraphics[width=\textwidth]{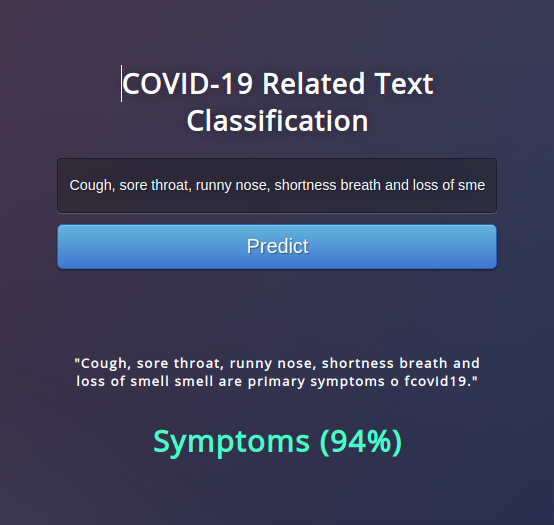}
        \caption{Demo of the web application}
    \label{website}
      \end{subfigure}
    
      \caption{Screenshot of chrome extension and real-time COVID-19 related text classification website}
    \end{figure}

As illustrated in Figure~\ref{extension}, the extension simplifies the process of accessing the classification website. When users click the extension, they are presented with the ``Go for classification" option, allowing them to swiftly access the website. Figure~\ref{website} shows the predicted class name along with its corresponding class accuracy. The deep learning model responsible for these predictions consistently achieves an impressive accuracy rate of 90.50\%.

\section{Discussion}\label{sec:Discussion}

The findings of this research hold significant implications for understanding and utilizing social media data in the context of public health, particularly during global health emergencies like the COVID-19 pandemic. The study's focus on classifying health-related terms within Twitter data contributes to the growing body of knowledge surrounding the role of social media in disseminating information and shaping public discourse during crises.

Numerous studies have concentrated on tweet classification related to COVID-19, covering a range of topics such as vaccine misinformation~\citep{WEINZIERL2021103955}, fake news~\citep{warman2023covidfakeexplainer}, stances toward online education~\citep{hamad2022steducov}, sentiment analysis~\citep{basiri2021novel} and emotion classification~\citep{oliveira2022investigating}.
The significance of this research lies in its attempt to bridge the gap in the existing literature~\citep{sanaullah2022applications} by specifically analysing COVID-19-related discussion on Twitter. By categorizing discussions into five distinct classes -- health risks, prevention, symptoms, transmission, and treatment -- the study provides a nuanced understanding of the diverse aspects of health-related conversations during the pandemic.
In the broader context of relevant studies, this research aligns with the trend of leveraging social media platforms for syndromic surveillance during global health emergencies, as seen in studies related to previous pandemics such as H1N1~\citep{signorini2011use}, Ebola~\citep{kim2016topic}, and SARS~\citep{zhu2008changes}. The findings complement existing literature by specifically addressing the dearth of studies focusing on health risks and transmission-related content in the context of COVID-19 on social media.

The empirical study's comparison of traditional machine learning and deep learning approaches is noteworthy~\citep{dargan2020survey}. The superior performance of the CNN algorithm, especially in comparison to traditional methods, highlights the potential of advanced techniques in extracting meaningful insights from social media data. This finding aligns with the evolving landscape of deep learning applications, emphasizing the importance of considering more sophisticated algorithms for analysing complex and dynamic datasets~\citep{barbieri2020tweeteval}.

The introduction of a new COVID-19 Twitter dataset is a practical contribution, enabling researchers and public health professionals to explore and analyse pandemic-related discussions comprehensively. The development of a web application prototype further underscores the practical applicability of the research, providing a tangible tool for real-time monitoring and analysis of health-related content on Twitter~\citep{morita2023tweeting,sinnenberg2017twitter}.

However, it is crucial to acknowledge the limitations inherent in this study. The exclusive training of the classifier on 140-character tweets raises questions about its adaptability to longer-form content and potentially affects prediction accuracy. Additionally, the focus on Twitter data may limit the generalizability of the findings to other social media platforms, as different platforms may exhibit distinct communication patterns and content structures. To address those limitations, one could consider expanding the training dataset to include longer-form content, allowing the classifier to adapt to diverse text lengths and potentially improving prediction accuracy across various content types~\citep{jones2022you}. Additionally, to enhance the generalizability of the findings, incorporating data from multiple social media platforms and adjusting the model to account for distinct communication patterns and content structures inherent to each platform would provide a more comprehensive understanding of health-related discussions in the broader digital landscape~\citep{khattar2022generalization}.

The challenges associated with accurately detecting health-related information highlight the complexities of analysing social media data for public health research. The misclassifications and nuances underscore the need for ongoing refinement and adaptation of advanced models to capture the subtleties of user-generated content accurately. In particular, the integration of large language models, such as GPT-3~\citep{brown2020language}, could offer a promising avenue for advancing the accuracy and efficiency of health-related content classification on social media~\citep{christensen2023prompt,tornberg2023chatgpt}. These models possess the capability to comprehend context, decipher nuanced language, and adapt to varying lengths of text, addressing some of the challenges associated with reported speech and the character limitations of tweets~\citep{kalyan2022ammu}. Leveraging such advanced language models in conjunction with the methodologies presented in this study could potentially enhance the classification accuracy and broaden the applicability of the system across diverse social media platforms and communication styles.
To further advance the field, future research should consider exploring the transferability of the model to other social media platforms. Additionally, comparative studies across different regions and demographic groups could provide valuable insights into the variations in health-related discussions on social media~\citep{li2018seeking}.

While this research makes significant contributions in analysing and classifying health-related discussions on Twitter during the COVID-19 pandemic, it also highlights the evolving nature of the digital landscape and the ongoing need for refinement and adaptation in methodologies. The findings contribute not only to the academic discourse but also offer practical tools for public health practitioners and policymakers to monitor and respond to health-related conversations in real-time. As the field continues to evolve, future studies should build upon these findings to enhance the effectiveness of utilizing social media data for public health surveillance and intervention strategies.

\section{Conclusion}\label{sec:Conclusion}
In this study, we harnessed machine learning algorithms to categorize health-related expressions within COVID-19 tweets. Our primary goal was to classify COVID-19 related tweets into five distinct classes: health risks, prevention, symptoms, transmission, and treatment. We curated a dataset comprising 6,667 tweets and meticulously annotated each one. This dataset underwent a comprehensive data refinement process, encompassing multiple data pre-processing steps. Additionally, we applied three distinct feature extraction techniques. Our study leveraged a combination of seven traditional machine learning algorithms, including Decision Tree, Random Forest, Stochastic Gradient Descent, K-nearest Neighbour, Adaboost, Logistic Regression, and Linear SVC, alongside four deep learning algorithms—LSTM, CNN, RNN, and BERT. Among the machine learning models, Stochastic Gradient Descent yielded the highest F1 score of 86.34\%, while the deep learning approach saw CNN delivering an impressive F1 score of 90\%.

The findings and analyses from this study signify that COVID-19 health-related phrases within prepared datasets can be effectively categorized using a spectrum of machine learning and deep learning algorithms. Given the distinct nature of our research, the proposed model could potentially serve as a standardized framework for the classification of COVID-19 health-related expressions within Twitter data. The outcomes of this research hold promise for global healthcare efforts against COVID-19 and offer valuable insights to researchers in this field.

However, it is important to acknowledge certain limitations within our study. We employed data from a limited timeframe and did not incorporate the entire available dataset. Our dataset relied on manual labeling, which restricted the volume of labeled data.

To address these limitations in future work, we intend to expand the dataset's size significantly. We are also exploring automated dataset labeling approaches to replace manual annotation. Furthermore, we plan to experiment with modified neural networks integrated with transfer learning techniques to enhance the study's robustness, accuracy, and overall outcomes.

\section*{Competing interests statement}
The authors declare that they have no known competing financial interests or personal relationships that could have appeared to influence the work reported in this paper.

\section*{Funding}
This research did not receive any specific grant from funding agencies in the public, commercial, or not-for-profit sectors.

 \bibliography{ref}





% \end{thebibliography}
\end{document}